\pdfoutput=1

\documentclass[11pt]{article}

\usepackage[]{ACL2023}

\usepackage{times}
\usepackage{latexsym}

\usepackage[T1]{fontenc}

\usepackage[utf8]{inputenc}

\usepackage{microtype}

\usepackage{inconsolata}

\usepackage{xcolor,soul}
\usepackage{booktabs}
\usepackage{pifont}
\usepackage{newunicodechar}
\usepackage{multirow}
\usepackage{amsmath}
\usepackage{svg}
\usepackage{nameref}
\usepackage{tabularx}

\title{Aligning AI Research with the Needs of Clinical Coding Workflows:\\Eight Recommendations Based on US Data Analysis and Critical Review}

\author{
 \textbf{Yidong Gan\textsuperscript{1, 2}},
 \textbf{Maciej Rybinski\textsuperscript{2, 3}},
 \textbf{Ben Hachey\textsuperscript{1, 4}},
 \textbf{Jonathan K. Kummerfeld\textsuperscript{1}}
\\
 \textsuperscript{1}The University of Sydney \hspace{0.25cm}
 \textsuperscript{2}CSIRO Data61 \hspace{0.25cm}
 \textsuperscript{3}The University of Málaga \hspace{0.25cm}
 \textsuperscript{4}Beamtree
\\
\texttt{yidong.gan@sydney.edu.au}
}

\begin{document}

\maketitle

\begin{abstract}
Clinical coding is crucial for healthcare billing and data analysis. Manual clinical coding is labour-intensive and error-prone, which has motivated research towards full automation of the process. However, our analysis, based on US English electronic health records and automated coding research using these records, shows that widely used evaluation methods are not aligned with real clinical contexts. For example, evaluations that focus on the top 50 most common codes are an oversimplification, as there are thousands of codes used in practice. This position paper aims to align AI coding research more closely with practical challenges of clinical coding. Based on our analysis, we offer eight specific recommendations, suggesting ways to improve current evaluation methods. Additionally, we propose new AI-based methods beyond automated coding, suggesting alternative approaches to assist clinical coders in their workflows.
\end{abstract}

\section{Introduction}
Clinical coding is a process that transforms clinical notes into a set of alphanumeric codes, which represent diagnoses and procedures during medical visits. This process is essential to tasks like hospital billing and disease prevalence studies. Manual clinical coding is labour-intensive and error-prone \cite{karimi-etal-2017-automatic, li2020icd}. To address these issues, automated coding has been widely explored. While many existing studies frame it as a multi-label classification task \cite{mullenbach-etal-2018-explainable, ijcai2020p0461, huang-etal-2022-plm, li2023towards}, no analysis has examined whether that matches the needs of clinicians.

This position paper critically reviews automated coding studies based on multi-label classification, focusing on those using the public US Medical Information Mart for Intensive Care (MIMIC) datasets \cite{goldberger2000physiobank, johnson2016mimic, johnson2023mimic}. For consistency, we use the clinical term `code' instead of `label' throughout this paper. We focus on the MIMIC datasets as they are far larger than other public datasets. Specifically, MIMIC provides English electronic health records of acute and emergent inpatients, with the latest version, MIMIC-IV, including 331,675 patient admissions. The few other public datasets, such as those released by \citet{pestian-etal-2007-shared} and \citet{miranda2020overview}, contain 978 and 1,000 admissions, respectively. This significant scale difference allows MIMIC to cover a broader range of diagnoses and procedures, making it more suitable for comprehensive analyses.

We show that current evaluations do not align with the needs of clinical contexts. In practice, coders must select from over a thousand codes and sequence them correctly \cite{guidline2024cm}, yet many studies use smaller code sets and overlook code sequencing in evaluation. Additionally, the Area Under the Receiver Operating Characteristic Curve (AUC-ROC) is often the only threshold-independent metric reported, which is inappropriate given the imbalanced code distribution. Common human coding metrics like Exact Match Ratio and Jaccard Score are often omitted, making it difficult to measure the accuracy gap between automated and human coding. In other words, widely used evaluation strategies do not measure what truly matters for clinical application.

We then propose new methodologies to support clinical coders rather than automating the entire process. Given the limited effectiveness of automated coding \cite{edin2023automated}, manual coding with software assistance remains prevalent, with code auditing essential in clinical workflows to reduce errors. Despite its significance, research on AI-assisted coding and auditing is limited. This underscores the need for future studies in these areas, which have the potential to yield practical solutions sooner than fully automated coding.

\section{Clinical Coding Workflow} \label{sec:workflow}
In this section, we present a typical clinical coding workflow in an inpatient setting, outline the individual coding processes, and provide an overview of the current state-of-the-art strategies and tools used in these processes. The workflow presented is specific to the US context; other countries may have similar or different processes.

\begin{figure}[t]
  \centering
  \includegraphics[width=1\columnwidth]{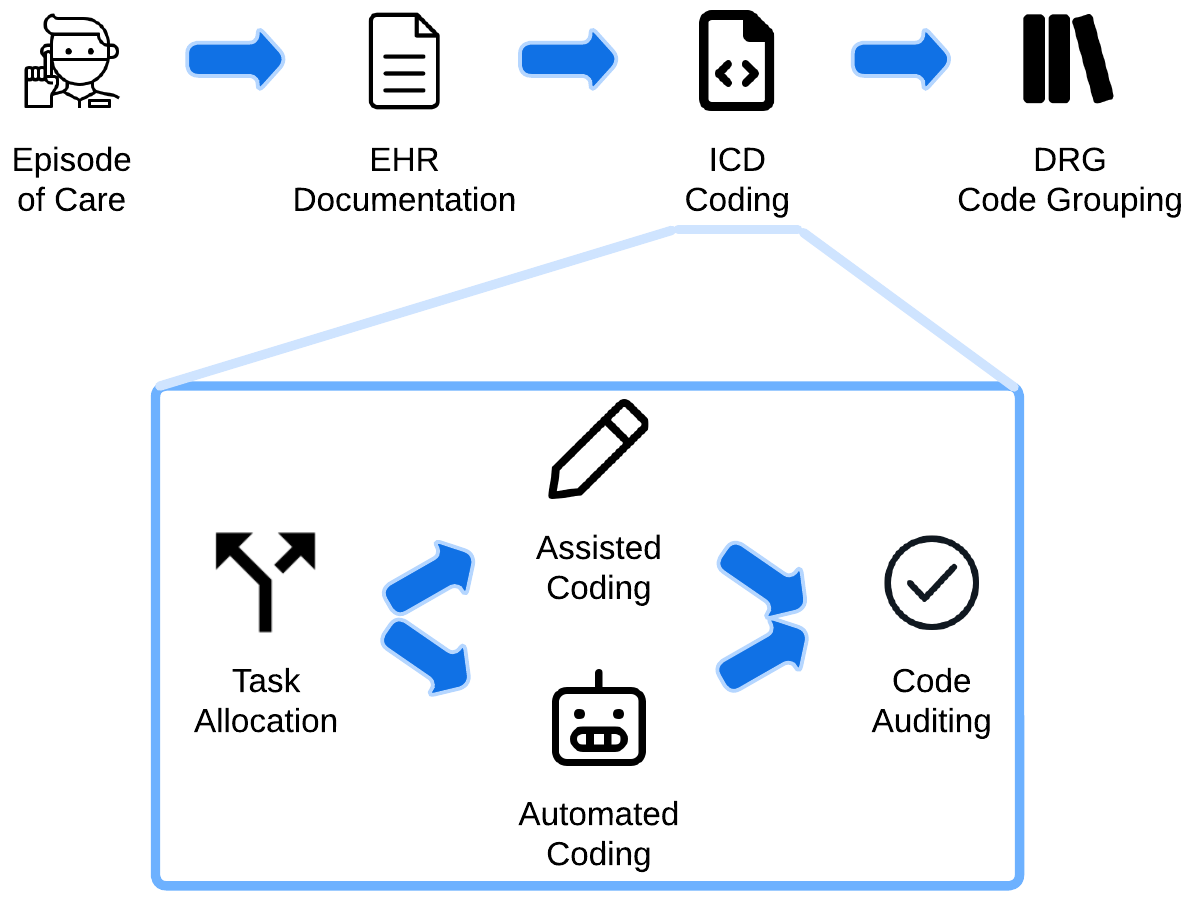}
  \caption{Typical clinical coding workflow for an inpatient episode of care in the US.}
  \label{fig:workflow}
\end{figure}

The upper diagram in Figure~\ref{fig:workflow} outlines a high-level, four-step clinical coding workflow for an inpatient episode. First, a patient is admitted to the hospital. During care, all relevant documentation (e.g., pathology reports) is entered into their Electronic Health Record (EHR). Upon discharge, the attending doctor writes a summary detailing the patient's stay, including diagnosis and treatment. Clinical coders then assign International Classification of Diseases (ICD) codes based on the EHR. For reimbursement purposes, relevant ICD codes are grouped into a Diagnosis Related Group (DRG) code. DRG is a classification system that organises hospital cases into groups; each DRG has a specific payment rate based on the average resources needed to treat patients in that group. The lower diagram in Figure \ref{fig:workflow} gives additional details of the ICD coding task, illustrating the processes before and after coding episodes with ICD.

The implementation of ICD and DRG varies across countries, and the coding inputs can differ depending on healthcare settings (e.g., hospital policies). Previous AI coding studies often use discharge summaries as inputs but have also explored other sources, such as radiology reports \cite{pestian-etal-2007-shared} and multilingual death certificates \cite{neveol2018clef}. Given all these differences, it is important that AI coding research carefully considers the clinical coding workflow it is addressing. Of note, many references in the following subsection are corporate products, as previous research has largely focused on the automated coding approach in the clinical coding workflow.

\subsection{Task Allocation}
With many hospitals facing backlogs of uncoded cases \cite{alonso2020problems}, optimising task allocation is crucial. The order in which cases are processed depends on hospital-specific business metrics, such as maximising profit and reducing backlog. Tools like Beamtree's Q Coding platform \cite{bmt2024qcoding} support task forecasting, rule-based distribution, and scheduling, enabling hospitals to address backlogs systematically and ensure critical cases are handled promptly.

After determining case priority, the next step is to allocate cases to the appropriate expert. In assisted coding (i.e., manual coding with computer software support), assigning cases based on coder specialty or experience effectively manages the complexity of coding tasks \cite{alonso2020problems, bmt2024qcoding}. Likewise, with AI-based coding solutions, cases could be assigned to either assisted or automated coding pathways. This step is speculative and will be discussed further in Section \ref{sec:analysis}.

\subsection{Assisted Coding}
In assisted coding, where coders use computer software to manually enter codes, each keystroke, word read, and thought incurs a cost. By minimising the manual effort involved, these tasks become more efficient. This promise of higher efficiency has motivated health tech companies to develop various assisted coding tools.

Existing tools include features such as searching and navigating codes with integrated guidelines, which help users quickly find relevant codes and follow best practices \cite{bmt2024qcoding, 3m2024codefinder}. AI-suggested codes and DRG grouping helps in streamlining the coding process \cite{bmt2024qcoding, 3m2024professional, 3m2024codefinder}. Customisable rules allow users to automate coding based on criteria drawn from their expertise \cite{3m2024professional}. Online audits against organisational policies ensure quality control \cite{3m2024auditexpert}. Evidence linking for assigned codes facilitates efficient edits and reviews \cite{goinvo2024coderyte, 3m2024codeassist}. Collectively, these features contribute to a more efficient and cost-effective coding workflow.

\subsection{Automated Coding}
Automated coding refers to the process of assigning accurate diagnostic and procedural ICD codes without human intervention. It is closely related to other real-world problems, such as tagging in social networks \cite{coope2019neural} and indexing biomedical literature \cite{krithara2023road}, where a piece of text is categorised using multiple labels. Automated coding can address coding of episodes where the accuracy of an AI system is higher than human performance or is otherwise sufficient given operational cost-benefit considerations. CodeAssist \cite{3m2024codeassist}, for example, is a commercial system used in many hospitals, with its main feature being automated coding.

Automated coding has also been the focus of AI coding research to date, with most studies relying solely on discharge summaries as model inputs. \citet{mullenbach-etal-2018-explainable} propose the Convolutional Attention for Multi-Label Classification (CAML) model. CAML combines convolutional neural networks (CNNs) with per-code attention to focus on text sections relevant to each ICD code. This attention mechanism enhances interpretability by highlighting the text that contributes to the model's decisions. \citet{li2020icd} extend this framework with the Multi-Filter Residual Convolutional Neural Network (MultiResCNN), which uses multiple filter layers and residual connections for better feature extraction. On the other hand, \citet{ijcai2020p0461} introduce the Label Attention Model (LAAT), using a bidirectional long short-term memory (LSTM) to capture clinical context in text. LAAT uses a distinct per-code attention mechanism compared to CAML, incorporating additional steps in attention weight calculations and a hierarchical joint learning strategy, which leads to better predictions for rare ICD codes. \citet{huang-etal-2022-plm} adapt LAAT's attention mechanism for pretrained language models in ICD coding (PLM-ICD). PLM-ICD also incorporates domain-specific pretraining and segment pooling, addressing challenges such as the domain mismatch between pretraining and clinical text and the large code space.

Recent benchmarking by \citet{edin2023automated} compares CAML, MultiResCNN, LAAT, PLM-ICD, and a few other models. PLM-ICD consistently achieves best results on MIMIC-III and MIMIC-IV datasets. However, all models, including PLM-ICD, struggle with rare codes, which is a persistent issue in automated coding. Notably, document length had minimal effect on model performance, with little difference when truncating documents from 4,000 to 2,500 words. In Section \ref{sec:analysis}, we demonstrate that prior studies fail to measure key factors critical for real-world applications. Our analysis reveals realistic upper bounds of existing state-of-the-art models compared to human performance.

\subsection{Code Auditing}
Clinical coding is complex, and even human coders often make mistakes \cite{burns2012systematic}. In the US, coding errors and quality improvement efforts cost an estimated \$25 billion annually \cite{xie-xing-2018-neural}. Even more concerning, some errors may be treated as fraud, leading to legal liability \cite{rudman2009healthcare}. To address this, many asynchronous auditing tools have been developed. For example, \citet{3m2024auditexpert} provides offline tools for batch auditing, referencing documents and codes used in patient claims to ensure compliant coding, and integrated denial tracking for managing coding quality. \citet{bmt2024qcoding} offers tools for audits against quality indicators, dynamic code sequencing, and code combination validation. In general, these tools aim to enhance coding quality and reduce error-related costs in both time and revenue.

\section{Data and Automation: Analysis and Recommendations} \label{sec:analysis}
In this section, we (1) identify key shortcomings in the evaluation methodologies widely used in automated coding studies, (2) analyse the widely used MIMIC datasets, and (3) offer corresponding recommendations. Many of these shortcomings are not limited to studies using the MIMIC datasets, but are also prevalent in other methodologies proposed for automated coding shared tasks involving different datasets \cite{pestian-etal-2007-shared, miranda2020overview}.

\begin{table*}[t]
\centering
\small
\begin{tabularx}{\textwidth}{XlX}
\toprule
 & Count & References \\
Reported only top 50 results & 8 &  \cite{shi2017towards, wang-etal-2018-joint-embedding, teng2020automatic, feucht-etal-2021-description, sun2021multitask, liu-etal-2022-treeman, michalopoulos-etal-2022-icdbigbird, liu2023chaticd} \\
Reported only full results & 2 & \cite{wiegreffe-etal-2019-clinical, kim2021read} \\
Reported one or more top k results, excluding full results & 3 & \cite{prakash2017condensed, xie-xing-2018-neural, zhang-etal-2020-bert} \\
Reported both top 50 and full results, but used top 50 results to emphasise contribution & 4 & \cite{li2020icd, li2021jlan, shi-etal-2021-analyzing, yang-etal-2022-knowledge-injected}  \\
Reported both top 50 and full results without emphasising top 50 results & 8 & \cite{mullenbach-etal-2018-explainable, xie2019ehr, ijcai2020p0461, cao-etal-2020-hypercore, liu-etal-2021-effective, zhou-etal-2021-automatic, yuan-etal-2022-code, li2023towards}  \\
\bottomrule
\end{tabularx}
\caption{Overview of Automated Coding Studies and Their Evaluation Strategy.}
\label{tab:top50_literature}
\end{table*}

\subsection{Evaluations Using the Top 50 Codes Do Not Reflect Real Effectiveness}
Table \ref{tab:top50_literature} shows that many of the existing studies evaluate their methods using the 50 most frequent codes\footnote{This is not intended as a systematic literature review. Instead, the aim is to demonstrate that many studies rely on the top 50 codes to validate their methodologies.}. When considering the application of models in a real-world healthcare environment, this evaluation strategy is sub-optimal \cite{liu-etal-2021-effective}; it focuses solely on the most frequent codes, failing to cover all diagnoses and procedures encountered in clinical practice.

\begin{figure}[t]
  \centering
  \includegraphics[width=1.0\columnwidth]{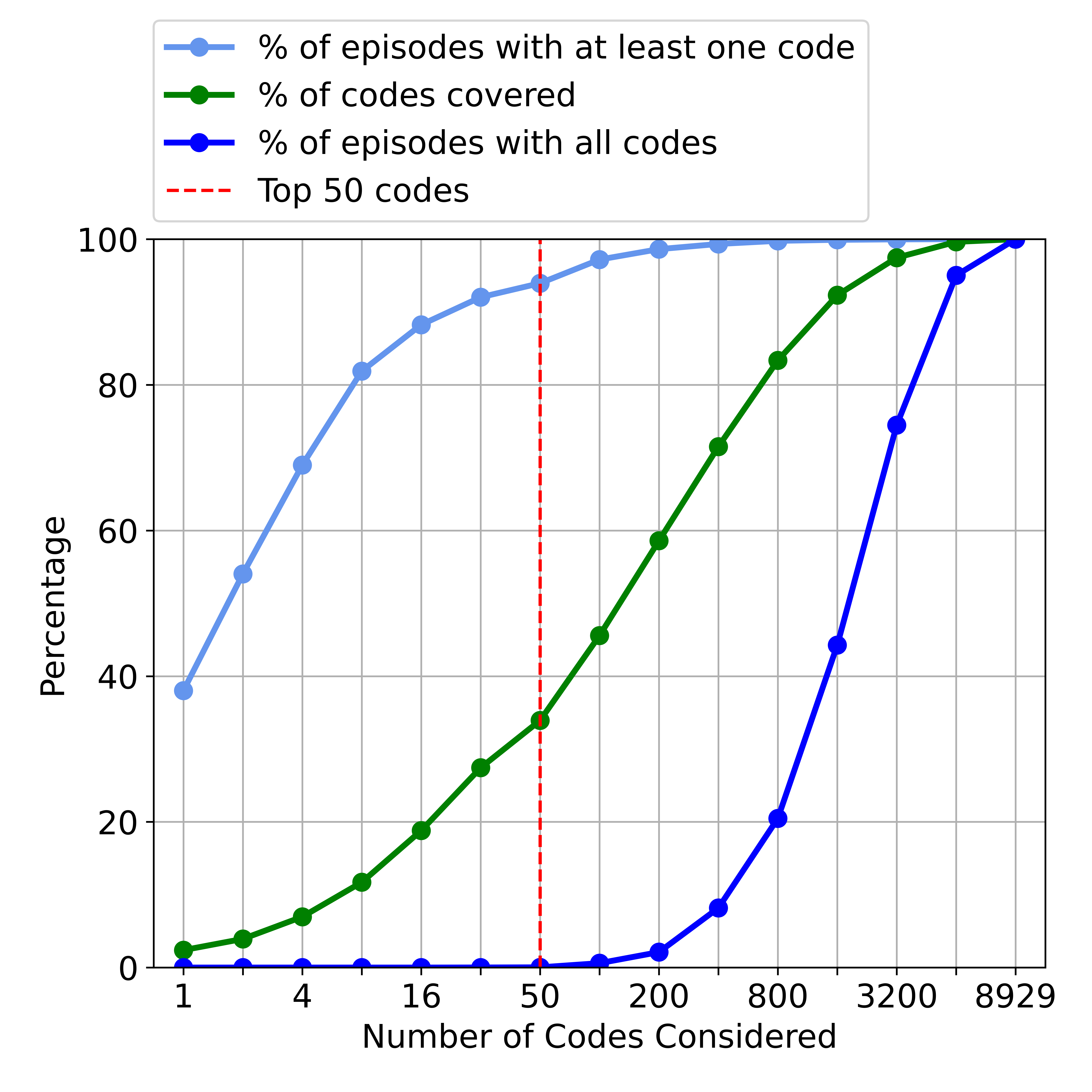}
  \caption{Code coverage in MIMIC-III.}
  \label{fig:top50}
\end{figure}

\begin{figure}[t]
  \centering
  \includegraphics[width=1.0\columnwidth]{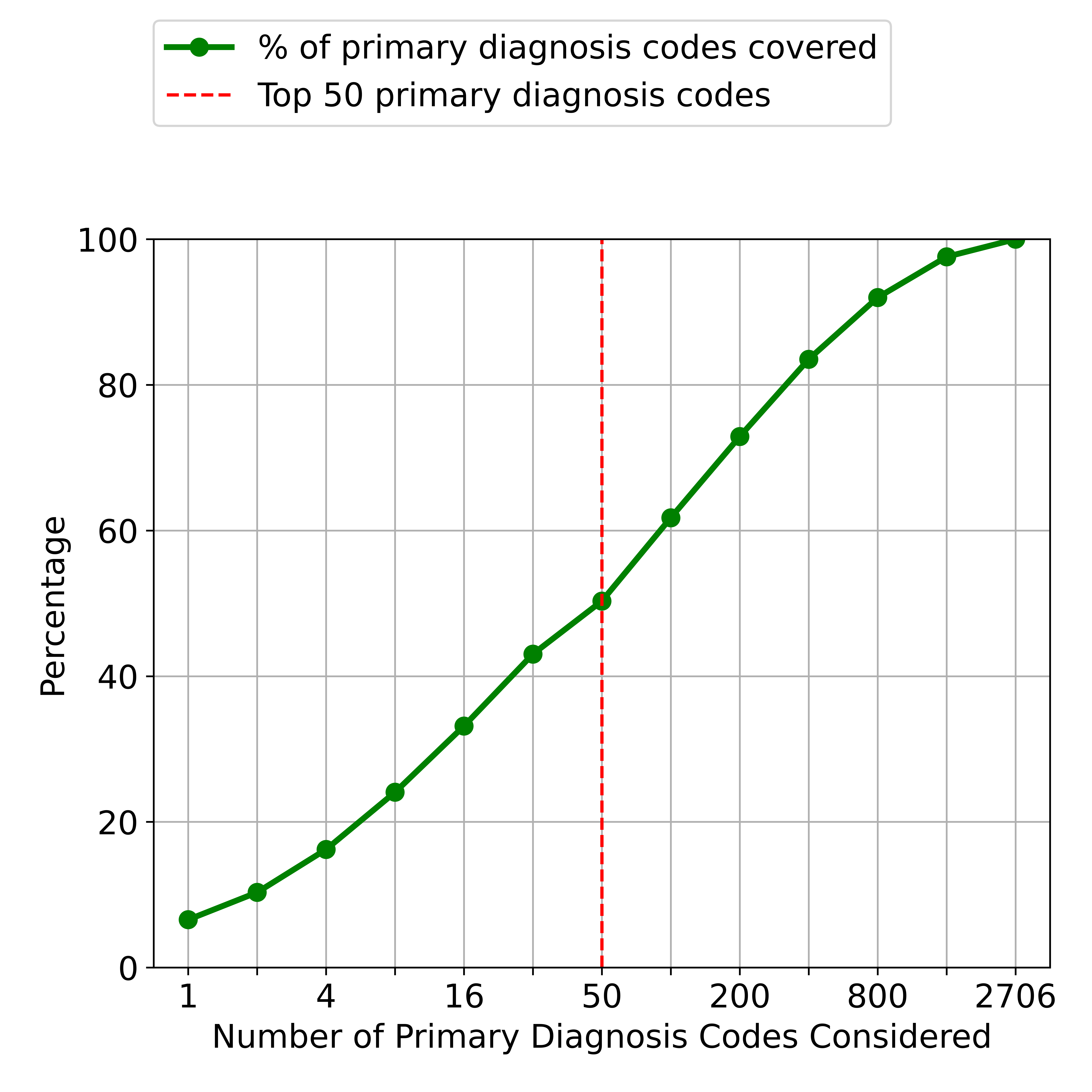}
  \caption{Code coverage for the first primary diagnosis code in an episode only.}
  \label{fig:top50_pdc}
\end{figure}

We demonstrate the coverage issue by showing the distribution of codes in MIMIC-III. The green line in Figure \ref{fig:top50} shows that the top 50 codes comprise only 33.92\% of total code occurrences. We further measure how well the episodes\footnote{A single episode usually has many codes.} in MIMIC-III are covered by the top 50 codes. Specifically, we calculate the percentage of episodes where all assigned codes are within the top 50. The dark blue line shows that none of the episodes are fully covered, meaning every episode has some codes outside the top 50. Even when we increase this to include the top 800 codes, the coverage rate remains very low, at 20.48\%. In other words, even with the top 800 codes, we miss some coding information in about 80\% of episodes. If we use a more generous measure, calculating the percentage of episodes that include at least one of the top 50 codes, the light blue line shows a coverage rate of 93.95\%. If we preprocess the MIMIC-III episodes to retain only the first primary diagnosis code, which typically has the highest clinical and billing priority, the issue of low coverage rates persists. We observe an 8\% higher coverage rate with the top 50 primary diagnosis codes (see Figure \ref{fig:top50_pdc}), yet the improvement is still marginal in terms of overall coverage.

The low coverage rate leads to a generalisation issue, where the ranking of different models shows low correlation between the top 50 and full code settings. According to \citet{mullenbach-etal-2018-explainable}, in the top 50 setting, CNN outperforms their proposed CAML model in terms of Micro F1, Macro F1, and Precision@5. However, CAML performs better in the full code setting. This contradiction is also observed in a reproducibility study \cite{edin2023automated}. If we rank the six reproduced models by macro AUC-ROC scores from lowest to highest, the order in the top 50 setting is Bi-GRU, CAML, CNN, MultiResCNN, LAAT, PLM-ICD, whereas in the full code setting, it is CNN, Bi-GRU, CAML, LAAT, MultiResCNN, PLM-ICD. \textbf{Therefore, we suggest that future studies compare models cautiously using top 50 results and prioritise full code setting improvements.}

\subsection{Global Thresholds Do Not Account for Variable Error Cost}
In many existing studies, a threshold is defined as the minimum confidence level required to assign a specific code, and a fixed threshold of 0.5 is used without justification \cite{mullenbach-etal-2018-explainable, wang-etal-2018-joint-embedding, li2020icd, ijcai2020p0461}. This is important because the standard evaluation includes the F1 score, a threshold-dependent classification metric. \citet{edin2023automated} found that not tuning the threshold significantly degraded the performance of most models, so they fine-tuned a single threshold by maximising the model's micro F1 score on the validation set. This does not account for the inherent differences between codes, such as prior probabilities and misclassification costs. This leads to an interesting direction of research: adapting dynamic thresholds \cite{wu-etal-2019-learning, alotaibi2021multi} for classification-based automated coding methods. \textbf{Additionally, this highlights the importance of threshold-independent metrics, such as Area Under the Receiver Operating Characteristic Curve (AUC-ROC) and Area Under the Precision-Recall Curve (AUC-PR), as they provide comprehensive assessment of automated coding models across various thresholds.}

\subsection{AUC-ROC Score Is Not Ideal for Imbalanced Datasets}
AUC-ROC is a binary classification metric. In multi-label classification, a common approach is to convert the task into a one-vs-all setting, where one code (class) is treated as the positive code and all remaining codes as the negative code. Then, the AUC-ROC scores for each code are computed individually. AUC-PR offers a different perspective by focusing on the relationship between precision and recall. One way to calculate AUC-PR is by using Average Precision (AP), which represents the average of precision values at different recall levels as the threshold varies. Since multiple codes can be predicted for each episode, we can calculate the mean AP for all possible codes, known as the Mean Average Precision (MAP). In other words, the AUC-PR for each code can be calculated using AP, while MAP extends this by averaging the AP values across all predicted codes.

In an imbalanced dataset like MIMIC, the dominance of the negative code can result in a misleadingly high AUC-ROC score. A study by \citet{edin2023automated} on three MIMIC splits shows that the SOTA automated coding model, PLM-ICD, consistently achieves macro AUC-ROC scores greater than 95\%, indicating it is very effective at scoring relevant codes higher than irrelevant ones. By only looking at this metric, one might infer that this is a robust model. However, PLM-ICD's MAP remains below 70\% across all three splits, indicating that when it predicts a certain code given various threshold values, it is often incorrect. Previous work \cite{mullenbach-etal-2018-explainable, liu-etal-2021-effective, yuan-etal-2022-code, huang-etal-2022-plm} reported only the AUC-ROC score, which can be misleading, as the model's precision trade-off is not well reflected. \textbf{Thus, we recommend reporting AUC-PR as well as AUC-ROC.}

\subsection{Automated Coding Evaluation Should Match Human Coding Evaluation}
\begin{table*}[t]
\centering
\small
\begin{tabularx}{0.85\textwidth}{lXccc}
\toprule
 &  & MIMIC-III Clean & MIMIC-IV ICD-9 & MIMIC-IV ICD-10 \\
\midrule
Three digits accuracy &  & 52.84 $\pm$ 0.34 & 55.22 $\pm$ 0.19 & 51.17 $\pm$ 0.22 \\
Four digits accuracy &  & 46.21 $\pm$ 0.33 & 49.28 $\pm$ 0.19 & 44.97 $\pm$ 0.22 \\
Full code accuracy &  & 44.01 $\pm$ 0.33 & 46.75 $\pm$ 0.18 & 42.05 $\pm$ 0.22 \\
\bottomrule
\end{tabularx}
\caption{The code accuracy of PLM-ICD calculated using Jaccard Score, with values spread within the 95\% confidence interval determined using the Z-score.}
\label{tab:label_acc}
\end{table*}

Existing studies aim for automated coding, which assumes that the model acts as an independent coder. However, their evaluations do not include the common accuracy metrics that are used to measure human performance.

The term `accuracy' can be confusing due to its many possible implementations. Different studies have reported human accuracy in clinical coding, but their definitions of accuracy are inconsistent \cite{burns2012systematic}. In this position paper, we will examine two implementations of accuracy. Instance accuracy, or Exact Match Rate (EMR), measures the percentage of instances (i.e., episodes of care or medical cases) where the predicted code set exactly matches the true code set. Code accuracy, or Jaccard Score, is defined as the ratio of the size intersection of a predicted code set and a target code set to the size of their union. Instance accuracy is stricter, as it requires a perfect match for every code in an instance.

In automated coding, even if a case contains a single error, human coders must re-code or correct the errors. This means the effort and cost associated with reviewing the entire clinical note are not mitigated, highlighting the need for measuring instance accuracy. On the other hand, code accuracy is useful because not all codes are equally important. When assigning a DRG code, the primary diagnosis code is usually the first and most important determinant factor \cite{drg2023au}. Given this, when evaluating coding accuracy, it is beneficial to allow for partial matches, acknowledging that capturing overlap in codes can still be valuable.

\textbf{We recommend that future studies include both of these accuracy metrics to better demonstrate the performance gap between AI and human coding.}

\subsection{Low Automation Accuracy Suggests Subset-Specific Automation} \label{sec:human_vs_model}
When considering instance accuracy, a study in the UK that included 50 episodes of care showed that human accuracy is 54\%. In a UK hospital setting, \citet{abdulla2020improving} reported an average of 67.5\%  accuracy each month over four months.  In contrast, Among the `MIMIC-III clean', `MIMIC-IV ICD-9', and `MIMIC-IV ICD-10' splits, none of the six advanced models replicated by \citet{edin2023automated} achieved an instance accuracy greater than 1.1\%.

When considering code accuracy, the median human performance in the UK was 83.2\%, with large variance among thirty-two studies (50-98\%) \cite{burns2012systematic}. However, the definition of accuracy is inconsistent across the explored studies; some defined inaccurate coding as inaccurate three digit coding, while the majority defined it as inaccurate four digit coding. In fact, many inaccuracies occur at the four digit level \cite{burns2012systematic} instead of the three digit level. We select PLM-ICD, the best model according to \citet{edin2023automated}, and report its code accuracy with respect to three-digit, four-digit, and full code levels across three MIMIC splits. For each split, we train a single PLM-ICD instance on the full code prediction task, adjusting the accuracy measure across different digits to account for varying evaluation complexities. Table \ref{tab:label_acc} shows that PLM-ICD's three digit accuracy, the most generous evaluation measure, is much higher than the other two in all splits. If we compare this result against human accuracy, PLM-ICD is only half as good as an average human at best, indicating there still exists a notable gap to reach full automation. 

Instead of full automation, we could consider the task allocation part of clinical coding workflows. A recent study in radiology \cite{agarwal2023combining} suggests that the best approach for combining human expertise with AI is to delegate cases to either AI or humans, rather than having AI augment human decisions. In other words, automating a subset of tractable episodes may be a promising direction for AI coding. The main challenge, however, is the choice of the subset. Figure \ref{fig:doc_unique_chapters} shows that in MIMIC-IV, approximately only 1\% of episodes contain one unique ICD-10-CM three-digit code, while more than half include at least six. This suggests we cannot simply choose the subset based on a single disease or symptom. \textbf{We encourage future research to investigate the selection of tractable subsets of care and estimate realistic upper bounds for these subsets.}

\subsection{MIMIC Episodes Are Challenging to Fully Automate}
\begin{figure}[t]
  \centering
  \includegraphics[width=1.0\columnwidth]{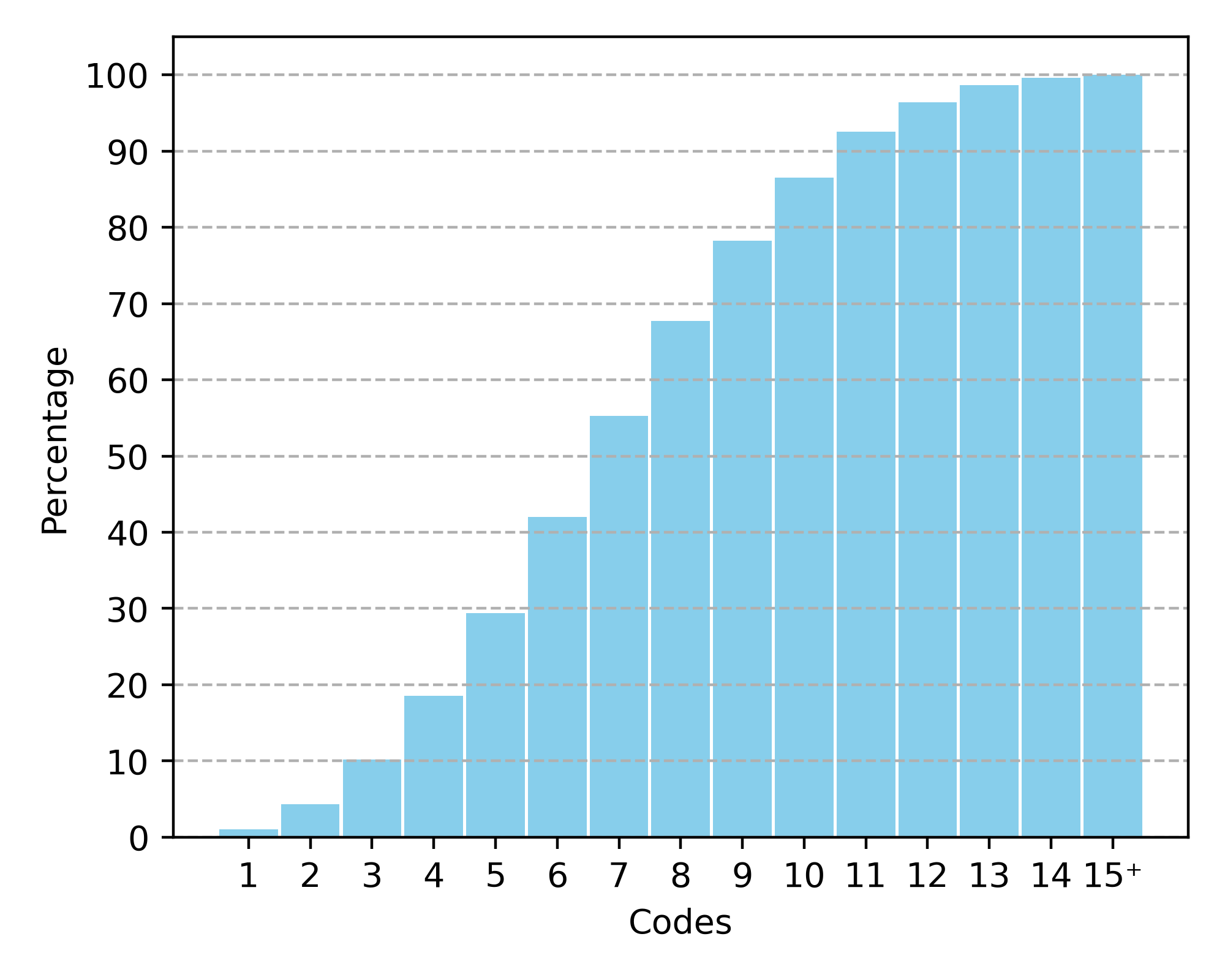}
  \caption{Distribution of MIMIC-IV episodes by the number of unique three-digit ICD-10-CM codes (e.g., G47 for 'sleep disorders'). About 4\% of the episodes have at most two of these three-digit codes.}
  \label{fig:doc_unique_chapters}
\end{figure}

The MIMIC cohort consists intensive care unit (ICU) and emergent inpatients, who often present complex conditions requiring multiple diagnoses (see Figure \ref{fig:doc_unique_chapters}) and treatments \cite{alonso2020problems}. More details on the MIMIC cohort can be found in Appendix \ref{sec:appendix-mimic}. \citet{campbell2020computer} noted that, compared to inpatient coders, outpatient coders are more concerned that assisted coding will replace their role. One possible reason for this is that outpatient episodes often involve less complex conditions \cite{alonso2020problems}. This suggests that outpatient episodes, which are not included in MIMIC, may be better automation candidates.

A common problem in the MIMIC datasets is the imbalanced code distribution, where less than half of the full codes occur at least 10 times, except in the MIMIC-IV ICD-9 collection (see Appendix \ref{sec:appendix-mimic-overview} for more details). MIMIC-III's coverage of the ICD-9-CM code space is relatively low, representing only 50.16\% of the possible 17,800 full codes \cite{wiki:ICD-10}. This limitation is worse in MIMIC-IV's ICD-10-CM/PCS data, which includes 18.78\% of the possible 139,000 full codes \cite{wiki:ICD-10}. In other words, the evaluations are omitting a large proportion of ICD codes that may be easier to automate but are missing or rare in MIMIC due to the nature of the dataset.

\textbf{Large public datasets that cover a broader range of care types are currently lacking. Therefore, developing such datasets would be invaluable for automated coding.}

\textbf{Additionally, we suggest exploring the use of MIMIC for AI coding research in other key parts of the workflow: task allocation, assisted coding, and code auditing. We will discuss this further in Section \ref{sec:methodology}.}

\subsection{Code Sequence Matters}
It is not sufficient to consider only the correctness of the assigned codes; their sequence is also important. For instance, certain conditions have both an underlying etiology and manifest across multiple body systems. The official ICD-10-CM American guideline requires the underlying condition to be sequenced before the manifestation \cite{guidline2024cm}. Similarly, the official Australian ICD-10-AM guideline requires the anaesthetic codes to be sequenced immediately following the procedure code to which they relate \cite{guideline2022am}. Therefore, it is evident that clinical coding requires the modeling of both code sequence and dependency. However, the sequence of the target codes is neglected by existing work despite this information being provided in the MIMIC datasets. In fact, both sequence and dependency issues are inherent in some multi-label classification tasks. Many corresponding solutions have been developed \cite{read2011classifier, alvares2012incorporating, yeh2017learning, yang2019hierarchical}. \textbf{We recommend that future studies include code sequences in their evaluation to better estimate the real impact on workflow.}

\section{New Workflow-Inspired Methodologies} \label{sec:methodology}
In Section \ref{sec:analysis}, we explored limitations of the current formulation of the task as a multi-label classification problem. Now, we propose alternatives to integrate AI into clinical coding workflows as recommendation systems or asynchronous (offline) auditing assistants. In these cases, information retrieval metrics such as Precision@k, Recall@k, and Coverage Error are more appropriate.

\subsection{New Assisted Coding Methodologies}
We consider three types of tasks to augment the manual coding process: (1) a sequential task, where the system predicts one code at a time and receives human feedback after each prediction; (2) a recall task, where coders start with a large set of possible codes in mind and select from a set of system suggested codes; and (3) a structural task, where coders take a top-down approach, and instead of predicting complete codes, the system only predicts partial codes.

We propose various assisted coding systems to address the three tasks. While these systems provide a great starting point, they are not exhaustive. The proposed assisted coding systems leverage both human and AI strengths, but their implementation is likely not straightforward. Multiple user designs can be implemented for the same system, leading to varying performance results. Additional challenges include the time needed for humans to adapt to collaborating with AI and the risk of over-trust or under-trust in AI suggestions \cite{agarwal2023combining}. Therefore, we recommend including user studies or field tests when evaluating assisted coding systems, measuring not only accuracy and efficiency but also user satisfaction and trust in the system, and user-AI effectiveness.

\subsubsection{The Sequential Task}
The goal of this task is to transform the multi-label classification problem into a simpler multi-classification problem. Instead of predicting all codes in a single step, the objective is to assign codes sequentially like real coding practice. We can implement this in several ways:
\begin{enumerate}
    \item Chain a group of classifiers sequentially, similar to the classifier chain proposed by \citet{read2011classifier} for related problems in other domains. Each classifier makes a binary choice about assigning a code and receives user feedback, where the feedback is then used as part of the input for the next classifier.
    \item Train a single multi-class classifier that predicts one code at a time, receives user feedback, and stops when a predefined termination criterion is met.
    \item Treat all ICD codes as vocabulary and use a seq2seq model to predict the codes, prompting the user for feedback after each decoding step.
\end{enumerate}

All three designs add sequential information as part of the input to tackle the code sequencing and dependency issue. Relevant evaluation metrics include Precision@k, the number of steps required to achieve full recall, and the model's convergence rate with human feedback. Importantly, while it is true that human-in-the-loop is not absolutely mandatory in the suggested designs, the inclusion of human input offers significant benefits. These include higher accuracy over long run, real-time error correction, and improved user trust (critical for clinical applications) as users contribute to the model's learning process.

\subsubsection{The Recall Task}
This task can be framed as a multiple-choice problem, where the objective is to maximise relevant choices while minimising the total number of choices. We propose two designs:

\begin{enumerate}
    \item A system where all predicted codes are presented to a human expert.
    \item A system where high-confidence predicted codes are assigned automatically, whereas low-confidence predicted codes are presented to a human expert.
\end{enumerate}

Both designs can still be approached using multi-label classification; however, the evaluation metrics and optimisation methods should differ. Typically, multi-label classification uses a loss function that penalises any incorrect outputs. In this task, it is acceptable if some of the model's top confident codes are incorrect, as long as the correct options are included. This approach involves optimising for Recall@k, ensuring high recall with a low value of k. In the second design, the ranking of positive codes becomes important as well. A key challenge in this task is designing effective choice minimisation strategies, which may include grouping by ontology or confidence interval. For example, grouping codes by related ICD categories can reduce the number of options a coder needs to review, thereby speeding up the coding process. 

\subsection{The Structural Task}
The structural task we propose is inspired by \citet{nguyen-etal-2023-two}, but their setup does not incorporate human input. The objective of their work is to use a two-stage decoder that first predicts the parent (i.e., the first three digits) codes and then the child (i.e., the digits following the first three) codes. Their model's parent prediction on `MIMIC-III Full' achieves micro and macro F1 scores of 29.1\% and 69.0\% respectively, outperforming the overall micro and macro F1 scores of 10.5\% and 58.4\% by a considerable margin. This confirms that parent prediction is much simpler for AI models. Based on this, we could design systems where the second stage benefit from human input:
\begin{enumerate}
    \item The system delegates the challenging task of predicting child codes entirely to humans, while its parent code predictions are used to augment human decision-making.
    \item The system could predict a set of parent codes, and upon a human selecting a parent code, it would use that input to predict the relevant child codes. This approach leverages human expertise to refine the AI system's broader categorisations.
\end{enumerate}

In addition to standard classification metrics, explainability metrics (e.g., matching annotated evidence spans) are useful for evaluating the first design, as the system output is intended to inform human decision-making. Hierarchical evaluation metrics (e.g., \citealp{falis-etal-2021-cophe}) are very useful as they account for the hierarchical code structure, avoiding equal penalisation for all mispredictions. Overall, the proposed designs are more robust in handling rare codes, as parent codes (i.e., the broader categories) have more training examples. By incorporating a human-in-the-loop approach, the system effectively reduces the likelihood of error propagation from the first stage, resulting in improved reliability and efficiency.

\subsection{New Code Auditing Methodologies}
Clinical coding is challenging and humans often make errors \cite{searle-etal-2020-experimental, cheng-etal-2023-mdace}. If a model achieves high Precision@k, it could be integrated as an offline auditor that starts up immediately after human coders finish coding an episode. For example, if a model's Precision@1 score is 95\%, then its most confident code is correct 95\% of the time. This is beneficial in addressing under-coding. For instance, the model can flag a high-confidence code that is missing in the episode, prompting the coder for review. Such offline designs ensure AI interventions do not disrupt the human's standard coding workflow, improving coding quality and reducing the back-and-forth communication between coders and auditors. When evaluating these models, considering specific metrics such as the decrease in under-coding incidents and the improvement in billing accuracy post-AI review will provide clear insights into the the AI's efficacy in improving the workflow.

\section{Summary of Recommendations}
\begin{enumerate}
    \item  The top 50 codes have low episode coverage, limiting the generalisation of evaluation results to the full code setting, which is more practical. We recommend against validating methodologies solely with the top 50 codes.
    \item Clinical coding involves variable error costs, different thresholds might be optimal for different codes. We recommend reporting more threshold-independent scores for a comprehensive evaluation.
    \item  In imbalanced datasets like MIMIC, a high AUC-ROC score does not necessarily indicate a high precision score due to the dominance of the negative code (class). We recommend reporting both AUC-PR and AUC-ROC.
    \item  Existing studies on automated coding often exclude common human accuracy metrics like EMR and Jaccard Score. We recommend including these metrics to demonstrate the practical effectiveness of automated solutions.
    \item Our findings indicate that current SOTA models do not achieve human accuracy on MIMIC datasets. We recommend more AI coding research on deliberate task allocation, focusing on a manageable subset of episodes.
    \item MIMIC includes acute inpatients, which are challenging even for human coders. We recommend creating datasets of other care types for automated coding research.
    \item MIMIC episodes are challenging to automate. We recommend using MIMIC datasets for AI coding research in other parts of clinical coding workflows, such as developing systems for code suggestion and coding audit assistance, as described in our proposed methodologies.
    \item  Code sequence is part of real-world coding practice. We recommend including it in future evaluations to provide a more comprehensive assessment of system effectiveness.
\end{enumerate}

\section{Conclusion}
Most studies have approached clinical coding as a traditional multi-label classification task. In this position paper, we conduct a critical review and data analysis of these studies and the MIMIC dataset. We show key shortcomings in existing evaluation methodologies, which fail to align with clinical contexts. We offer eight recommendations for aligning research with the needs of clinical coding workflows. Additionally, we introduce new AI-based methodologies beyond automated coding, proposing alternative research directions with a practical impact on clinical coding workflows.

\section{Limitations}
Our analysis focuses exclusively on automated coding studies that use the US MIMIC datasets. Therefore, some findings, such as the coverage issue of the top 50 common codes, may not generalise to datasets with patient cohorts different from MIMIC. Likewise, workflow-related discussions may not be universally applicable. This limitation, however, arises from the scarcity of public clinical coding datasets, which is also why MIMIC datasets are widely used in AI coding research. One of our eight recommendations is to create new public datasets for other types of care, which would benefit the entire AI coding community and enable more comprehensive evaluations.

\section*{Acknowledgements}

This material is partially supported by the Australian Research Council through a Discovery Early Career Researcher Award and the Commonwealth Scientific and Industrial Research Organisation (CSIRO).
We also thank the anonymous reviewers for their constructive feedback on our submission.

\bibliography{anthology,custom}
\bibliographystyle{acl_natbib}

\appendix
\section{AI Coding Task Definition} \label{sec:appendix-definition}
Previous studies have approached clinical coding as a fine-grained multi-label classification task. The objective is to create a model that maps an input text (usually a discharge summary) to a set of labels (ICD codes). The term `fine-grained' indicates that the code space is extremely large, covering thousands of codes or more. Both code frequency and the number of associated codes per instance (an episode or medical case) is highly imbalanced. The code space is presented as a hierarchy, where codes exhibit parent-child relationships.

\section{Descriptive Statistics of the MIMIC Cohort} \label{sec:appendix-mimic}
An overview of the patient cohorts in MIMIC-III/IV is shown in Table \ref{tab:mimic-cohort}. The preprocess code is built upon the work of \citet{edin2023automated}. Admissions without assigned ICD codes are removed. Age is calculated based on the patients' age at the time of admission. The Elixhauser index is calculated using \cite{vc2023comorbidipy}, with \citet{quan2005coding}’s mapping and \citet{van2009modification}’s score weighting, and is adjusted for patient age. LOS refers to the length of stay. Some admissions contain missing or invalid data, such as missing LOS values or cases where the discharge time is earlier than the admission time. Note that many ICU LOS values are missing in MIMIC-IV. These admissions are excluded from corresponding statistical calculations, and the number of included admissions, $N$, is provided accordingly.

\begin{table*}[t]
\small
\centering
\begin{tabularx}{\textwidth}{lXcccc}
\toprule
&  & \multicolumn{2}{c}{ICD-9} &  & ICD-10 \\
\cmidrule{3-4} \cmidrule{6-6} 
&  & MIMIC-III & MIMIC-IV &  & MIMIC-IV \\
\midrule
Years collected &  & 2001--2012 & 2008--2019 &  & 2008--2019 \\
Number of admissions &  & 52,722 & 209,359 &  & 122,316 \\
Number of patients &  & 41,126 & 97,727 &  & 65,685 \\
Age: Median (IQR) &  & 63 (49--77) & 61 (48--74) &  & 61 (48--72) \\
Elixhauser index: Median (IQR) &  & 7 (2--14) & 6 (2--12) &  & 7 (2--15) \\
\multirow{2}{*}{LOS: Median (IQR) [days]} & & 7.18 (4.26--12.78) & 3.28 (1.79--5.96) & & 3.88 (2.02--6.96) \\
& & \textit{N=52,671} & \textit{N=209,306} & & \textit{N=122,297} \\
\multirow{2}{*}{ICU LOS: Median (IQR) [days]} & & 2.43 (1.30--5.29) & 1.95 (1.09--3.84) & & 2.14 (1.18--4.30) \\
& & \textit{N=51,938} & \textit{N=38,717} & & \textit{N=26,606} \\
\bottomrule
\end{tabularx}
\caption{Overview of the MIMIC-III v1.4 and MIMIC-IV v2.2 patient cohort.}
\label{tab:mimic-cohort}
\end{table*}

\begin{table*}[t]
\centering
\small
\begin{tabularx}{\textwidth}{lXcccc}
\toprule
&  & \multicolumn{2}{c}{ICD-9} &  & ICD-10 \\ \cmidrule{3-4} \cmidrule{6-6} 
&  & MIMIC-III & MIMIC-IV &  & MIMIC-IV \\
\midrule
Words per document: Median (IQR) &  & 1,375 (965--1,900) & 1,320 (997--1,715) &  & 1,492 (1,147--1,931) \\ 
Full codes per document: Median (IQR) &  & 14 (10--20) & 12 (8--17) &  & 15 (10--21) \\ 
Number of unique three-digit codes &  & 1,606 & 1,712 &  & 2,239 \\
Number of unique four-digit codes &  & 6,120 & 7,337 &  & 11,254 \\
Number of unique full codes &  & 8,929 & 11,331 &  & 26,098 \\
Full code coverage [\%] &  & 50.16 & 63.66 &  & 18.78 \\
Full code freq $\geq$ 10 [\%] &  & 41.23 & 54.28 &  & 30.43 \\
\bottomrule
\end{tabularx}%
\caption{Overview of the MIMIC-III v1.4 and MIMIC-IV v2.2 ICD coding data. The preprocess code is built upon the work of \citet{edin2023automated}. Only admissions without assigned ICD codes are removed.}
\label{tab:mimic-overview}
\end{table*}

\section{Overview of the MIMIC datasets} \label{sec:appendix-mimic-overview}
Table \ref{tab:mimic-overview} shows descriptive statistics of the ICD coding data in MIMIC-III/IV. In the ICD-10 collection of MIMIC-IV, the median document length is 1,492 words with an interquartile range (IQR) of 1,147-1,931, and the median number of full codes per summary is 15, with an IQR of 10-21. These are comparable across the other two collections. The varying number of codes in most summaries reflects considerable differences in the complexity of medical issues documented, aligning with the fluctuating Elixhauser index presented in Table \ref{tab:mimic-cohort}.
\end{document}